\documentclass{article}

\usepackage[square,sort,comma,numbers]{natbib}

\usepackage[final]{report}

\usepackage[utf8]{inputenc} 
\usepackage[T1]{fontenc}    
\usepackage{hyperref}       
\usepackage{url}            
\usepackage{booktabs}       
\usepackage{amsfonts}       
\usepackage{nicefrac}       
\usepackage{microtype}      
\usepackage{bm}
\usepackage{amsmath}
\usepackage{algorithm}
\usepackage{algorithmic}
\usepackage{subfigure}
\usepackage{wrapfig}

\newcommand{\xv}{\bm{x}}

\newcommand{\hv}{\bm{h}}

\newcommand{\thetav}{\bm{\theta}}

\newcommand{\data}{\mathcal{D}}

\newcommand{\yv}{\bm{y}}

\newcommand{\wv}{\bm{w}}

\newcommand{\mmd}{\textrm{MMD}}

\newcommand{\cmmd}{\textrm{CMMD}}

\newtheorem{theorem}{\bf{Theorem}}
\newtheorem{remark}{\bf{Remark}}

\usepackage{color}
\newcommand{\junz}[1]{{\color{red}{\bf\sf[#1]}}}

\usepackage{tikz}
\usetikzlibrary{fit,positioning,arrows,decorations.markings}
\newcommand{\drawrec}[5]{
	\draw[fill=#5,thick] (#1,#2) rectangle (#3,#4);
}
\tikzstyle{vecArrow} = [thick, decoration={markings,mark=at position
   1 with {\arrow[semithick]{open triangle 60}}},
   double distance=1.4pt, shorten >= 5.5pt,
   preaction = {decorate},
   postaction = {draw,line width=1.4pt, white,shorten >= 4.5pt}]
\tikzstyle{innerWhite} = [semithick, white,line width=1.4pt, shorten >= 4.5pt]
\title{Conditional Generative Moment-Matching Networks}

\author{
  Yong Ren,$\ \ $ Jialian Li,$\ \ $ Yucen Luo,$\ \ $ Jun Zhu \\
  Department of Computer Science, Tsinghua University \\
  \texttt{\{renyong15@mails, luoyc15@mails, jl12@mails, dcszj@mail\}.tsinghua.edu.cn} \\
}

\begin{document}

\maketitle

\begin{abstract}
  Maximum mean discrepancy (MMD) has been successfully applied to learn deep generative models for characterizing a joint distribution of variables via kernel mean embedding.
In this paper, we present conditional generative moment-matching networks (CGMMN),
which learn a conditional distribution given some input variables
based on a conditional maximum mean discrepancy (CMMD) criterion.
The learning is performed by stochastic gradient descent with the gradient calculated by back-propagation.
We evaluate CGMMN on a wide range of tasks, including predictive
modeling, contextual generation, and Bayesian dark knowledge, which distills knowledge from a
Bayesian model by learning a relatively small CGMMN student network.
Our results demonstrate competitive performance in all the tasks.
\end{abstract}

\section{Introduction}
Deep generative models (DGMs) characterize the distribution of observations with a multilayered structure of hidden variables under nonlinear transformations. Among various deep learning methods, DGMs are natural choice for those tasks that require probabilistic reasoning and uncertainty estimation, such as image generation~\citep{LapGAN:nips15}, multimodal learning~\citep{multimodalDBM}, and missing data imputation. Recently, the predictive power, which was often shown inferior to pure recognition networks (e.g., deep convolutional networks), has also been significantly improved by employing the discriminative max-margin learning~\citep{MMDGM:nips15}.

For the arguably more challenging unsupervised learning, \cite{GAN:nips2014} presents a generative adversarial network (GAN),
which adopts a game-theoretical min-max optimization formalism. GAN has been extended with success in various tasks~\citep{Mirza:arxiv14,LapGAN:nips15}. However, the min-max formalism is often hard to solve.
The recent work~\citep{GMMD:icml2015,MMD:uai2015} presents generative moment matching networks (GMMN), which has a simpler objective function than GAN while retaining the advantages of deep learning. GMMN defines a generative model by sampling from some simple distribution (e.g., uniform) followed through a parametric deep network. To learn the parameters, GMMN adopts maximum mean discrepancy (MMD)~\citep{KTST:jmlr}, a moment matching criterion where kernel mean embedding techniques are used to avoid unnecessary assumptions of the distributions. Back-propagation can be used to calculate the gradient as long as the kernel function is smooth.

A GMMN network estimates the joint distribution of a set of variables. However, we are more interested in a conditional distribution in many cases, including (1)
{\it predictive modeling}: compared to a generative model that defines the joint distribution $p(\xv, \yv)$ of input data $\xv$ and response variable $\yv$,
	a conditional model $p(\yv | \xv)$ is often more direct without unnecessary assumptions on modeling $\xv$,
    and leads to better performance with fewer training examples~\citep{disc-vs-gen,crf2001};
(2) {\it contextual generation}: in some cases, we are interested in generating
samples based on some context, such as class labels~\citep{Mirza:arxiv14}, visual attributes~\citep{attribute2image} or the input information in cross-modal generation (e.g., from image to text \cite{image2text} or vice versa \cite{text2image});
and (3) {\it building large networks}: conditional distributions are essential building blocks of a large generative
probabilistic model. One recent relevant work~\citep{LapGAN:nips15} provides a good example of stacking multiple conditional GAN networks~\citep{Mirza:arxiv14} in a Laplacian pyramid structure to generate natural images.

In this paper, we present conditional generative moment-matching networks (CGMMN) to learn a flexible conditional distribution when some input variables are given.
CGMMN largely extends the capability of GMMN to address
a wide range of application problems as mentioned above,
while keeping the training process simple.
Specifically, CGMMN admits a simple generative process, which
draws a sample from a simple distribution and then passes the sample as well as the given conditional variables through a deep network to generate a target sample.
To learn the parameters, we develop conditional maximum mean discrepancy (CMMD),
which measures the Hilbert-Schmidt norm (generalized Frobeniu norm)
between the kernel mean embedding of an empirical conditional distribution and that of our generative model. Thanks to the simplicity of the conditional generative model, we can easily draw a set of samples to estimate the kernel mean embedding as well as the CMMD objective. Then, optimizing the objective can be efficiently implemented via back-propagation.
%
%
We evaluate CGMMN in a wide range of tasks, including predictive modeling, contextual generation, and Bayesian dark knowledge~\cite{BDK:arXiv}, an interesting case of
distilling dark knowledge from Bayesian models.
Our results on various datasets demonstrate that CGMMN can obtain competitive performance in all these tasks.

\vspace{-.3cm}
\section{Preliminary}
\vspace{-.3cm}

In this section, we briefly review some preliminary knowledge, including maximum mean
discrepancy (MMD) and kernel embedding of conditional distributions.

\vspace{-.3cm}
\subsection{Hilbert Space Embedding}
\vspace{-.2cm}
We begin by providing an overview of Hilbert space embedding, where we represent distributions
by elements in a 
{\it reproducing kernel Hilbert space} (RKHS).
A RKHS $\mathcal{F}$ on $\mathcal{X}$ with kernel $k$ is a Hilbert space of functions $f:\mathcal{X} \to \mathbb{R}$.
Its inner product $\langle \cdot, \cdot \rangle_{\mathcal{F}}$ satisfies the reproducing property:
$
\langle f(\cdot), k(\xv,\cdot) \rangle_{\mathcal{F}} = f(\xv).
$
Kernel functions are not restricted on $\mathbb{R}^d$. They can also be defined on graphs, time series
and structured objects~\cite{kernelML2008}.
We usually view $\phi(\xv) := k(\xv,\cdot)$ as a (usually infinite dimension) feature map of $\xv$. The most
interesting part is that we can embed a distribution by taking expectation on its feature map:
$$
\mu_X := \mathbb{E}_X[\phi(X)] = \int_{\Omega} \phi(X) dP(X).
$$
If $\mathbb{E}_X[k(X,X)] \leq \infty $, $\mu_X$ is guaranteed to be an element in the RKHS.
This kind of kernel mean embedding provides us another perspective on manipulating distributions whose parametric forms are not
assumed, as long as we can draw samples from them. This technique has been widely
applied in many tasks, including feature extractor, density estimation and two-sample test~\cite{KED:icalt2007,KTST:jmlr}.

\vspace{-.1cm}

\vspace{-.1cm}
\subsection{Maximum Mean Discrepancy}
\vspace{-.1cm}

Let $X= \{\xv_i\}_{i=1}^N$ and $Y = \{\yv_i\}_{j=1}^M$ be the sets of samples from distributions $P_X$ and $P_Y$, respectively.
Maximum Mean Discrepancy (MMD), also known as kernel two sample test~\cite{KTST:jmlr}, is a frequentist
estimator to answer the query whether $P_X = P_Y$ based on the observed samples.
The basic idea behind MMD is that if the generating distributions are identical, all the
statistics are the same. Formally, MMD defines the following difference measure: 
$$
	\mmd[\mathcal{K},P_X,P_Y] := \sup\limits_{f\in\mathcal{K}}( \mathbb{E}_X[f(X)] - \mathbb{E}_Y[f(Y)]),
$$
where $\mathcal{K}$ is a class of functions. \cite{KTST:jmlr} found that the class of
	functions in an universal RKHS $\mathcal{F}$ is rich enough to
	distinguish any two distributions and MMD can be expressed as the difference of their mean embeddings.
	Here, universality requires that $k(\cdot,\cdot)$ is continuous and $\mathcal{F}$ is dense in
	$C(X)$ with respect to the $L_{\infty}$ norm, where $C(X)$ is the space of bounded continuous functions on $X$.
	We summarize the result in the following theorem:
    \begin{theorem}{\cite{KTST:jmlr}}
\label{thm1}
	Let $\mathcal{K}$ be a unit ball in a universal RKHS $\mathcal{F}$, defined on the compact
	metric space $\mathcal{X}$, with an associated continuous kernel $k(\cdot,\cdot)$. When the
	mean embedding $\mu_p,\mu_q \in \mathcal{F}$, the MMD objective function can be expressed as
	$\mmd[\mathcal{K},p,q] = \|\mu_p - \mu_q\|^2_{\mathcal{F}}$.
	Besides, $\mmd[\mathcal{K},p,q] = 0$ if and only if $p=q$.
\end{theorem}
In practice, an estimate of the MMD objective compares the square difference between the empirical kernel mean embeddings:
\begin{equation*}
\begin{aligned}
    \widehat{\mathcal{L}}_{\mmd}^2	 &= \left \|\dfrac{1}{N}\sum\limits_{i=1}^{N} \phi(\xv_i)
- \dfrac{1}{M}\sum\limits_{j=1}^{M}\phi(\yv_i) \right \|^2_{\mathcal{F}} , 
\end{aligned}
\end{equation*}
which can be easily evaluated by expanding the square and using the associated kernel $k(\cdot,\cdot)$. Asymptotically, $\widehat{\mathcal L}^2_{\mmd}$ is an unbiased estimator.

\vspace{-.1cm}
\subsection{Kernel Embedding of Conditional Distributions}
\label{kecd}
\vspace{-.1cm}

The kernel embedding of a conditional distribution $P(Y|X)$ is defined as:
$
\mu_{Y|\xv} := \mathbb{E}_{Y|\xv}[\phi(Y)] = \int_{\Omega}\phi(\yv)dP(\yv|\xv).
$
Unlike the embedding of a single distribution, 
the embedding of a conditional distribution
is not a single element in RKHS, but sweeps out a family of points in the RKHS, each
indexed by a fixed value of $\xv$.
Formally, the embedding of a conditional distribution is represented as
an operator $C_{Y|X}$, which satisfies the following properties:
\begin{equation}
\begin{aligned}
	1.\ \ \mu_{Y|\xv} = C_{Y|X}\phi(\xv); \ \ \ \	2.\ \ \mathbb{E}_{Y|\xv}[g(Y)|\xv] =  \langle g,\mu_{Y|\xv}\rangle_{\mathcal{G}},
\end{aligned}
\end{equation}
where $\mathcal{G}$ is the RKHS corresponding to $Y$.

\cite{HSECD:icml09} found that such an operator exists under some assumptions, using the technique
of cross-covariance operator
$C_{XY}: \mathcal{G} \to \mathcal{F}$:
	$$C_{XY} := \mathbb{E}_{XY}[\phi(X) \otimes \phi(Y)] - \mu_X \otimes \mu_Y,$$	
where $\otimes$ is the tensor product. An interesting property is that $C_{XY}$ can
also be viewed as an element in the tensor product space $\mathcal{G} \otimes \mathcal{F}$.
The result is summarized as follows.
\begin{theorem}{\cite{HSECD:icml09}}
	Assuming that $\mathbb{E}_{Y|X}[g(Y)|X] \in \mathcal{F}$, the embedding of conditional distributions
	$C_{Y|X}$ defined as $C_{Y|X} := \mathcal{C}_{YX}\mathcal{C}^{-1}_{XX}$ satisfies properties
	$1$ and $2$.
\end{theorem}

Given a dataset $\data_{XY} = \{(\xv_i,\yv_i)\}_{i=1}^N$ of size $N$ drawn $i.i.d.$ from
$P(X,Y)$, we can estimate the conditional embedding operator as
$\widehat{C}_{Y|X} = \Phi(K+\lambda I)^{-1}\Upsilon^{\top}$, where
$\Phi = (\phi(\yv_1),...,\phi(\yv_N)) , \Upsilon= (\phi(\xv_1),...,\phi(\xv_N)), K = \Upsilon^{\top}\Upsilon$
and $\lambda$ serves as
regularization.
The estimator is an element in the tensor product space $\mathcal{F} \otimes \mathcal{G}$ and
satisfies properties $1$ and $2$ asymptotically. When the domain of $X$ is finite, we can also
estimate $C_{XX}^{-1}$ and $C_{YX}$ directly (See Appendix~\ref{discrete} for more details).

\vspace{-.2cm}
\section{Conditional Generative Moment-Matching Networks}
\vspace{-.3cm}

We now present CGMMN, including a conditional maximum mean discrepancy criterion as the training objective, a deep generative architecture and a learning algorithm.

\vspace{-.2cm}
\subsection{Conditional Maximum Mean Discrepancy}
\vspace{-.2cm}

Given conditional distributions $P_{Y|X}$ and $P_{Z|X}$, we aim to test whether they are the same
in the sense that when $X = \xv$ is fixed whether $P_{Y|\xv} = P_{Z|\xv}$ holds or not.
When the domain of $X$ is finite, a straightforward solution is to test whether $P_{Y|\xv} = P_{Z|\xv}$ for each
$\xv$ separately by using MMD.
However, this is impossible when $X$ is continuous. Even in the finite case, as the separate tests do not share statistics,
we may need an extremely large number of training data to test a different model for each single value of $\xv$.
Below, we present a conditional maximum mean discrepancy criterion, which avoids the above issues.

Recall the definition of kernel mean embedding of conditional distributions. When $X=\xv$ is fixed, we
have the kernel mean embedding $\mu_{Y|\xv} = C_{Y|X}\phi(\xv)$. As a result, if we have
$C_{Y|X} = C_{Z|X}$, then $\mu_{Y|\xv} = \mu_{Z|\xv}$ is also satisfied for every fixed $\xv$. By the
virtue of Theorem \ref{thm1}, that $P_{Y|\xv} = P_{Z|\xv}$ follows as the following theorem states.
\begin{theorem}
Assuming that $\mathcal{F}$ is a universal RKHS with an associated kernel $k(\cdot,\cdot)$,
 $\mathbb{E}_{Y|X}[g(Y)|X] \in \mathcal{F}$,
 $\mathbb{E}_{Z|X}[g(Z)|X] \in \mathcal{F}$ and $C_{Y|X},~ C_{Z|X} \in \mathcal{F} \otimes \mathcal{G}$. If the embedding of
conditional distributions $C_{Y|X} = C_{Z|X}$, then $P_{Y|X} = P_{Z|X}$ in the sense that
for every fixed $\xv$, we have $P_{Y|\xv} = P_{Z|\xv}$.
\end{theorem}
The above theorem gives us a sufficient condition to guarantee that two conditional distributions
are the same. We use the operators to measure the difference of two conditional distributions
and we call it {\it conditional maximum mean discrepancy} (CMMD), which is defined as follows:
$$
\mathcal{L}_{\cmmd}^2 = \left \| C_{Y|X} - C_{Z|X}\right \|^2_{\mathcal{F} \otimes \mathcal{G}}.
$$
Suppose we have two sample sets $\mathcal{D}^s_{XY} = \{(\xv_i,\yv_i)\}_{i=1}^N$ and $\mathcal{D}^d_{XY} = \{(\xv_i,\yv_i)\}_{i=1}^M$.
Similar as in MMD, in practice 
we compare the square difference between the empirical estimates of the conditional embedding operators:
\begin{equation*}
\begin{aligned}
	\widehat{\mathcal{L}}_{\cmmd}^2= & \left \| \widehat{C}^d_{Y|X} - \widehat{C}^s_{Y|X}  \right \|_{\mathcal{F}\otimes \mathcal{G}}^2 , 
\end{aligned}
\end{equation*}
where the superscripts $s$ and $d$ denote the two sets of samples, respectively. For notation clarity, we define $\widetilde{K} = K + \lambda I$. Then, using kernel tricks, we can compute the difference only in term of kernel gram matrices:
\begin{equation}\label{eq:CMMD-Obj}
\begin{aligned}
	\widehat{\mathcal{L}}_{\cmmd}^2 	=&	\left \| \Phi_d(K_d + \lambda I)^{-1}\Upsilon_d^{\top}  - \Phi_s(K_s + \lambda I)^{-1}\Upsilon_s^{\top}\right \|_{\mathcal{F} \otimes \mathcal{G}}^2  \\
=	& \mathrm{Tr}\left( K_d \widetilde{K}_d^{-1} L_d \widetilde{K}_d^{-1} \right)   +	  \mathrm{Tr}\left( K_s \widetilde{K}_s^{-1} L_s \widetilde{K}_s^{-1} \right)
 -  2\cdot\mathrm{Tr}\left( K_{sd} \widetilde{K}_d^{-1} L_{ds} \widetilde{K}_s^{-1} \right),
\end{aligned}
\end{equation}
where $\Phi_d := (\phi(\yv^d_1), ..., \phi(\yv^d_N))$  and $\Upsilon_d := (\phi(\xv^d_1),..., \phi(\xv^d_N))$ are implicitly formed feature matrices, $\Phi_s$ and $\Upsilon_s$ are defined similarly for dataset $\mathcal{D}^s_{XY}$.
$K_d = \Upsilon^{\top}_d \Upsilon_d$ and $K_s = \Upsilon_s^{\top}\Upsilon_s$ are the gram matrices for input variables, while
$L_d = \Phi^{\top}_d \Phi_d$ and $L_s = \Phi_s^{\top}\Phi_s$ are the gram matrices for output variables. Finally, $K_{sd} = \Upsilon^{\top}_s \Upsilon_d$ and $L_{ds} = \Phi_d^{\top} \Phi_s$ are the gram matrices between the two datasets on input and out variables, respectively.

It is worth mentioning that we have assumed that the conditional mean embedding operator
	$C_{Y|X} \in \mathcal{F} \otimes \mathcal{G}$ to have the CMMD objective well-defined, which needs some smoothness assumptions
    such that $C_{XX}^{-3/2}C_{XY}$ is Hilbert-Schmidt~\cite{CMMAR:icml2012}.
	In practice, the assumptions may not hold, however, the
	empirical estimator $\Phi(K+\lambda I)^{-1} \Upsilon^{\top}$ is always an element
	in the tensor product space which gives as a well-justified approximation (i.e., the Hilbert-Schmidt norm exists) for practical use~\cite{HSECD:icml09}.

\begin{remark}
	Taking a close look on the objectives of MMD and CMMD, we can find some interesting
	connections.
	Suppose $N=M$. By omitting the constant scalar, the objective function of MMD can be rewritten as
	$$ \widehat{\mathcal{L}}_{\mmd}^2 = \mathrm{Tr}(L_d \cdot \bm{1}) + \mathrm{Tr}(L_s \cdot \bm{1}) -
	2 \cdot \mathrm{Tr}(L_{ds} \cdot \bm{1}), $$ where $\bm{1}$ is the matrix with all entities equaling to $1$.
	The objective function of CMMD can be expressed as
	$$ \widehat{\mathcal{L}}_{\cmmd}^2 = \mathrm{Tr}(L_d \cdot C_1) + \mathrm{Tr}(L_s \cdot C_2) -
	2 \cdot \mathrm{Tr}(L_{ds} \cdot C_3), $$
	where $C_1, C_2, C_3$ are some matrices based on the
	conditional variables $\xv$ in both data sets. 
	The difference is that instead of putting uniform weights on the gram matrix as in MMD,
	CMMD applies non-uniform weights, reflecting the influence of conditional variables. Similar observations
    have been shown in~\cite{HSECD:icml09} for the conditional mean operator, where the estimated conditional embedding $\mu_{Y|\xv}$ is
	a non-uniform weighted combination of $\phi(\xv_i)$. 
\end{remark}

\vspace{-.3cm}
\subsection{CGMMN Nets}
\vspace{-.3cm}

We now present a conditional DGM and train it by the CMMD  criterion.
One desirable property of the DGM is that we can easily draw samples
from it to estimate the CMMD objective. Below, we present such a network
that takes both the given conditional variables and an extra set of random variables as inputs,
and then passes through a deep neural network with nonlinear transformations to produce the samples
of the target variables.

\begin{wrapfigure}{r}{0.4\textwidth}\vspace{-.45cm}
\centering
\includegraphics[width=0.38\textwidth]{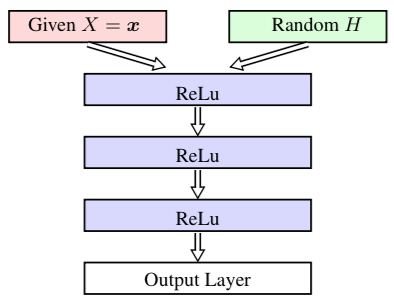}\vspace{-.35cm}
\caption{An example architecture of CGMMN networks. }
\label{fig1}\vspace{-.4cm}
\end{wrapfigure}

Specifically, our network is built on the fact that
for any distribution $\mathcal{P}$ on sample space $\mathbb{K}$ and any continuous distribution $Q$ on $\mathbb{L}$ that are
regular enough, there is a function $G: \mathbb{L} \to \mathbb{K}$ such that $G(\xv) \sim \mathcal{P}$, where
$\xv \sim \mathcal{Q}$~\cite{Prob}. This fact has been recently explored by~\cite{MMD:uai2015, GMMD:icml2015} to define a deep generative model
and estimate the parameters by the MMD criterion.
For a conditional model, we would like the function $G$ to depend on the given values of input variables.
This can be fulfilled via a process as illustrated in
Fig.~\ref{fig1}, where the inputs of a deep neural network (DNN)  
consist of two parts --- the input variables $\xv$ and an extra set of stochastic
variables $H \in \mathbb{R}^d$ that follow some continuous distribution.
For simplicity, we put a uniform prior on each hidden unit
$
p(\bm{h}) = \prod\limits_{i=1}^{d} U(h_i),
$
where $U(h) = \bm{I}_{(0\leq h \leq 1)}$
is a uniform distribution on $[0,1]$ and $\bm{I}_{(\cdot)}$ is
the indicator function that equals to $1$ if the predicate holds and $0$ otherwise.
After passing both $\xv$ and $\hv$ through the DNN,
we get a sample from the conditional distribution $P(Y|\xv)$:
$\bm{y} = f( \bm{x}, \bm{h} | \bm{w}),$
where $f$ denotes the deterministic mapping function represented by the network with parameters $\wv$.
By default, we concatenate $\bm{x}$ and $\bm{h}$ and fill $\widetilde{\bm{x}} = (\bm{x}, \bm{h})$ into the network.
In this case, we have $\bm{y} = f(\widetilde{\bm{x}} | \bm{w})$.

Due to the flexibility and rich capability of DNN on fitting nonlinear functions, this generative process can characterize
various conditional distributions well.
For example, a simple network can consist of
multiple layer perceptrons (MLP) activated by some non-linear functions such as the rectified linear
unit (ReLu)~\cite{Relu:icml2010}.
Of course the hidden layer is not
restricted to MLP, as long as it supports gradient propagation. We also use convolutional neural networks (CNN) as
hidden layers~\cite{CNN-SVHN:ICPR2012} in our experiments.
It is worth mentioning that there exist other ways to combine the conditional
variables $\xv$ with the auxiliary variables $H$. For example, we can add a corruption noise to the conditional
variables $\bm{x}$ to produce the input of the network, e.g., define $\widetilde{\bm{x}} =  \bm{x} + \bm{h}$, where $\bm{h}$ may follow a Gaussian distribution $\mathcal{N}(0, \eta I)$ in this case.

With the above generative process, we can train the network by optimizing the CMMD objective with proper regularization.
Specifically, let $\data_{XY}^s = \{(\xv_i^d, \yv_i^d)\}_{i=1}^{N}$ denote the given training dataset.
To estimate the CMMD objective, we draw a set of samples from the above generative model, where the conditional variables can be set by sampling from the training set with/without small perturbation (More details are in the experimental section).
Thanks to its simplicity, the sampling procedure can be easily performed. Precisely,
we provide each $\xv$ in the training dataset to the generator to get a new sample
and we denote $\data_{XY}^d = \{(\xv_i^s, \yv_i^s)\}_{i=1}^{M}$ as the generated samples.
Then, we can optimize the CMMD objective in Eq.~(\ref{eq:CMMD-Obj}) by gradient descent. See more
details in Appendix~\ref{grad_cal}.


\vspace{-.3cm}
\begin{algorithm}[ht!]
\caption{Stochastic gradient descent for CGMMN}
\label{alg1}
\centering
\begin{algorithmic}[1]
	\STATE {\bfseries Input:} Dataset $\mathcal{D} = \{(\bm{x}_i,\yv_i)\}_{i=1}^{N}$
	\STATE {\bfseries Output:} Learned parameters $\bm{w}$
    \STATE Randomly divide training dataset $\mathcal{D}$ into mini batches
	\WHILE{Stopping criterion not met}
	\STATE Draw a minibatch $\mathcal B$ from $\data$; 
	\STATE For each $\xv \in \mathcal B$, generate a $\yv$; and set $\mathcal B^\prime$ to contain all the generated $(\xv, \yv)$; 
	\STATE Compute the gradient $\frac{\partial \widehat{\mathcal{L}}^2_{\cmmd}}{\partial \bm{w}}$ on $\mathcal B$ and $\mathcal B^\prime$;
	\STATE Update $\bm{w}$ using the gradient with proper regularizer.
	\ENDWHILE
\end{algorithmic}
\end{algorithm}
\vspace{-.1cm}

Note that the inverse matrices $\widetilde{K}_s^{-1}$ and $\widetilde{K}_d^{-1}$ in the CMMD objective 
are independent of the model parameters, suggesting that we are not restricted to use differentiable kernels
on the conditional variables $\xv$.
Since the computation cost for kernel gram matrix grows cubically with the sample size,
we present an mini-batch version algorithm in Alg.~\ref{alg1} and some
discussions can be found in Appendix~\ref{mini-train}.

\vspace{-.4cm}
\section{Experiments}
\vspace{-.4cm}

We now present a diverse range of applications to evaluate our model, including
predictive modeling, contextual generation and an interesting case of Bayesian dark knowledge~\cite{BDK:arXiv}. Our results demonstrate that CGMMN is competitive in all the tasks.

\vspace{-.3cm}
\subsection{Predictive Performance}
\vspace{-.1cm}

\vspace{-.1cm}
\subsubsection{Results on MNIST dataset}
\vspace{-.3cm}

We first present the prediction performance on the widely used MINIST dataset, which consists of images in $10$
classes. Each image is of size $28 \times 28$ and the gray-scale is normalized to be in range $[0,1]$. The
whole dataset is divided into $3$ parts with $50,000$
training examples, $10,000$ validation examples and $10,000$ testing examples.

For prediction task, the conditional variables are the images $\bm{x} \in [0,1]^{28\times 28}$, and the generated
sample is a class label, which is represented as a vector $\bm{y} \in  \mathbb{R}_+^{10}$ and each $y_i$ denotes the
confidence that $\bm{x}$ is in class $i$. We consider two types of architectures in CGMMN --- MLP and
CNN.

\begin{wraptable}{r}{0.43\textwidth}
\vspace{-.5cm}
\caption{Error rates (\%) on MNIST dataset}
\label{table1}
\centering
\begin{tabular}{lc}
	\hline
	\hline
	Model  &  Error Rate \\
	\hline
    VA+Pegasos~\cite{MMDGM:nips15} & 1.04\\
    MMVA~\cite{MMDGM:nips15} & 0.90\\
	CGMMN & 0.97 \\
	\hline
    CVA + Pegasos~\cite{MMDGM:nips15} & 1.35 \\
	CGMMN-CNN & 0.47 \\
	\hline
    Stochastic Pooling~\cite{SP:iclr2013} & 0.47 \\
    Network in Network~\cite{NIN:iclr2014} & 0.47 \\
    Maxout Network~\cite{MN:icml2013} & 0.45 \\
    CMMVA~\cite{MMDGM:nips15} & 0.45 \\
    DSN~\cite{DSN:aistats2015} & 0.39 \\
	\hline
\end{tabular}
\vspace{-.3cm}
\end{wraptable}

We compare our model, denoted as CGMMN in the MLP case and CGMMN-CNN in the CNN case,
with Varitional Auto-encoder (VA)~\cite{VA:iclr2014},
which is an unsupervised DGM learnt by stochastic variational methods. To use VA for classification, a subsequent classifier is built --- We first learn feature representations by VA and then learn a linear SVM on these features using
Pegasos algorithm~\cite{Pegasos}.
We also compare with max-margin DGMs (denoted as MMVA with MLP as hidden layers and CMMVA
in the CNN case)~\cite{MMDGM:nips15}, which is a state-of-the-art DGM for prediction, and several other strong baselines, including Stochastic Pooling~\cite{SP:iclr2013},
Network in Network~\cite{NIN:iclr2014}, Maxout Network~\cite{MN:icml2013} and Deeply-supervised nets (DSN)~\cite{DSN:aistats2015}.

In the MLP case, the model architecture is shown in Fig.~\ref{fig1} with an uniform
distribution for hidden variables of dimension $5$. Note that since we do not need much randomness for
the prediction task, this low-dimensional hidden space is sufficient. In fact, we did not observe much
difference with a higher dimension (e.g., 20 or 50), which simply makes the training slower. 
The MLP has $3$ hidden layers with hidden unit number $(500,200,100)$ with the ReLu activation function. A minibatch size of $500$ is adopted.
In the CNN case, we use the same architecture as~\cite{MMDGM:nips15}, where there are $32$ feature maps
in the first two convolutional layers and $64$ feature maps in the last three hidden layers. An MLP of
$500$ hidden units is adopted at the end of convolutional layers. The ReLu activation function is used in the
convoluational layers and sigmoid function in the last layer. We do not pre-train our model and
a minibatch size of $500$ is adopted as well. The total
number of parameters in the network is comparable with the competitors~\cite{MMDGM:nips15, DSN:aistats2015, NIN:iclr2014, MN:icml2013}.

In both settings, we use AdaM~\cite{Adam:iclr2015} to optimize parameters.
After training, we simply draw a sample from our model conditioned on the input image
and choose the index of  maximum element of $\bm{y}$ as its prediction.Table \ref{table1} shows the results. We can see that
CGMMN-CNN is competitive with various state-of-the-art competitors that do not use data
augumentation or multiple model voting (e.g., CMMVA). DSN benefits from using more supervision signal in every
hidden layer and outperforms the other competitors.

\vspace{-.4cm}
\subsubsection{Results on SVHN dataset}
\vspace{-.3cm}

\begin{wraptable}{r}{0.42\textwidth}\vspace{-1.4cm}
\caption{Error rates (\%) on SVHN dataset}
\label{table3}
\centering
\begin{tabular}{lc}
	\hline
	\hline
	Model  &  Error Rate \\
	\hline
    CVA+Pegasos~\cite{MMDGM:nips15} & 25.3\\
	CGMMN-CNN & 3.13 \\
	\hline
	\hline
    CNN~\cite{CNN-SVHN:ICPR2012} & 4.9 \\
    CMMVA~\cite{MMDGM:nips15} & 3.09 \\
    Stochastic Pooling~\cite{SP:iclr2013} & 2.80 \\
    Network in Network~\cite{NIN:iclr2014} & 2.47 \\
    Maxout Network~\cite{MN:icml2013} & 2.35 \\
    DSN~\cite{DSN:aistats2015} & 1.92 \\
	\hline
\end{tabular}\vspace{-.5cm}
\end{wraptable}

We then report the prediction performance on the Street View House Numbers (SVHN) dataset. SVHN is a large dataset consisting of color images of size $32 \times 32$ in 10 classes. 
The dataset consists of $598,388$ training examples, $6,000$ validation examples and $26,032$ testing examples.
The task is significantly harder than classifying hand-written digits. Following~\cite{CNN-SVHN:ICPR2012,MMDGM:nips15}, we preprocess the data by Local Contrast Normalization (LCN). The architecture
of out network is similar to that in MNIST and we only use CNN as middle layers here. A minibatch size of $300$ is
used and the other settings are the same as the MNIST experiments.

Table \ref{table3} shows the results. Through there is a gap between our CGMMN
 and some discriminative deep networks such as DSN, our results are comparable with
those of CMMVA, which the state-of-the-art DGM for prediction. CGMMN is compatible with
various network architectures and we are expected to get better results with more sophisticated structures.

\vspace{-.2cm}
\subsection{Generative Performance}
\vspace{-.3cm}

\subsubsection{Results on MNIST dataset}
\vspace{-.2cm}

We first test the generating performance on the widely used MNIST dataset.
For generating task, the conditional variables are the image labels. Since $\yv$ takes a finite number of values, as mentioned in
Sec.~\ref{kecd}, we estimate $C_{YX}$ and $C_{XX}^{-1}$ directly and combine them as the estimation
of $C_{Y|X}$ (See Appendix~\ref{discrete} for practical details).
\begin{figure}[ht!]\vspace{-.2cm}
\centering
\subfigure[MNIST samples]{\includegraphics[width=4.4cm]{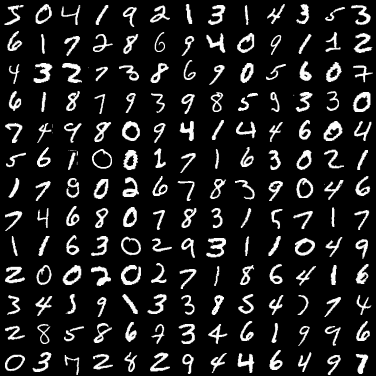}}
\subfigure[Random CGMMN samples]{\includegraphics[width=4.4cm]{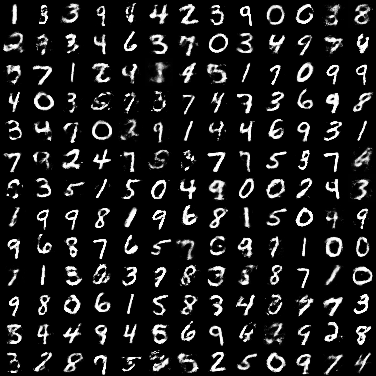}}
\subfigure[Samples conditioned on label $0$]{\includegraphics[width=4.4cm]{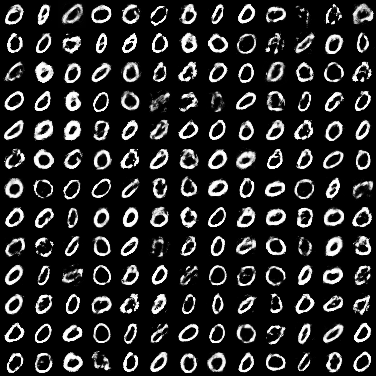}}\vspace{-.3cm}
\caption{Samples in (a) are from MNIST dataset; (b) are generated randomly from our CGMMN network;
(c) are generated randomly from CGMMN with conditions on label $y=0$. Both (b) and (c) are generated after running
$500$ epoches.}
\label{gen_fig}\vspace{-.45cm}
\end{figure}

The architecture is the same as before but exchanging the position of $\xv$ and $\yv$. For the
input layer,
besides the label information $\yv$ as conditional variables (represented by a one-hot-spot vector of dimension $10$),
we further draw a sample from a uniform distribution of dimension $20$, which is sufficiently large. Overall, the network is
a $5$-layer MLP with input dimension $30$ and the middle layer hidden unit number $(64,256,256,512)$, and
the output layer is of dimension $28 \times 28$, which represents the image in pixel.
A minibatch of size $200$ is adopted.

Fig.~\ref{gen_fig} shows some samples generated using our CGMMN, where in (b) the conditional variable $\yv$ is randomly
chosen from the 10 possible values, and in (c) $\yv$ is pre-fixed at class $0$.
As we can see, when conditioned on label $0$, almost all the
generated samples are really in that class.

\begin{wrapfigure}{r}{0.45\textwidth}\vspace{-.4cm}
\centering
\includegraphics[width=5.4cm]{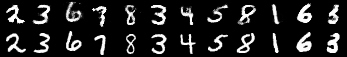}\vspace{-.3cm}
\caption{CGMMN samples and their nearest neighbour in MNIST dataset. The first row is our
generated samples.}
\label{fig_sim}\vspace{-.6cm}
\end{wrapfigure}

As in~\cite{GMMD:icml2015}, we investigate whether the models learn to merely copy the data. We
visualize the nearest neighbors in the MNIST dataset of several samples generated by CGMMN
in terms of Euclidean pixel-wise distance~\cite{GAN:nips2014} in Fig.~\ref{fig_sim}. As we can see, by this metric,
the samples
are not merely the copy.

\begin{wrapfigure}{r}{0.3\textwidth}\vspace{-.6cm}
\centering
\includegraphics[width=4.2cm]{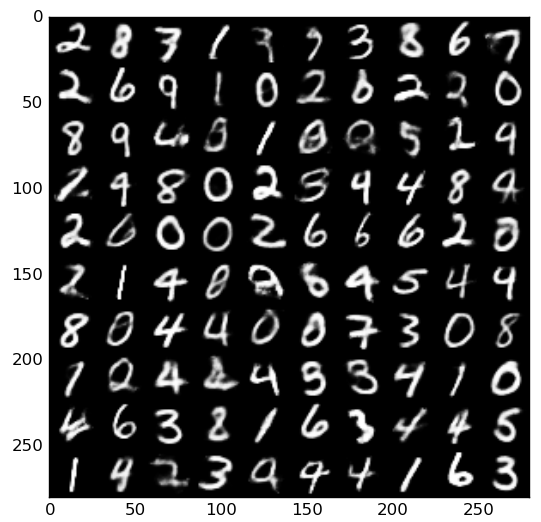}\vspace{-.6cm}
\caption{Samples generated by CGMMN+Autoencoder. The architecture follows from~\citep{GMMD:icml2015}.}
\label{cgmmn_auto}\vspace{-1cm}
\end{wrapfigure}

As also discussed in~\cite{GMMD:icml2015},
real-world data can be complicated and high-dimensional
and autoencoder can be good
at representing data in a code space that captures enough
statistical information to reliably reconstruct the data.
For example, visual data, while represented in a high dimension often exists
on a low-dimensional manifold.
Thus it is beneficial to combine
autoencoders with our CGMMN models to generate more smooth images,
in contrast to Fig.~\ref{gen_fig} where there are some noise in the generated samples.
Precisely, we first learn an auto-encoder and produce code representations of the
training data, then freeze the auto-encoder weights and learn
a CGMMN to minimize the CMMD objective between the generated codes using our CGMMN and
the training data codes. 
The generating results are shown in Fig.~\ref{cgmmn_auto}.
Comparing to Fig.~\ref{gen_fig}, the samples are more clear.\\

\vspace{-.5cm}
\subsubsection{Results on Yale Face dataset}
\vspace{-.2cm}
\label{yale_exp}
We now show the generating results on the Extended
Yale Face dataset~\cite{ext_yale_face}, which contains $2,414$ grayscale
images for $38$ individuals of dimension $32 \times 32$. 
There are about $64$ images per subject, one per
different facial expression or configuration.
A smaller version of the dataset consists of
$165$ images of $15$ individuals and the generating result can be found in
Appendix~\ref{yale-small}.

We adopt the same architecture as the first generating experiment for MNIST, which
is a $5$-layer MLP with an input dimension of $50$ ($12$ hidden variables and $38$ dimensions for
conditional variables, i.e., labels) and the middle layer hidden unit number
$(64, 256, 256, 512)$. A mini-batch size of $400$ is adopted. The other settings are the same
as in the MNIST experiment. The
overall generating results are shown in Fig.~\ref{fig_yale_ext}, where
we really generate diverse images for different individuals. Again, as shown in Appendix~\ref{yale-inter},
the generated samples are not merely the copy of training data.

\begin{wrapfigure}{r}{0.36\textwidth}\vspace{-.5cm}
\centering
\includegraphics[width=4.5cm]{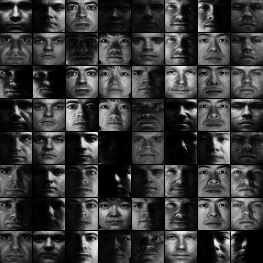}
\caption{CGMMN generated samples for Extended Yale Face Dataset. Columns are conditioned on different individuals.}
\label{fig_yale_ext}\vspace{-0.3cm}
\end{wrapfigure}

\vspace{-.3cm}
\subsection{Distill Bayesian Models}
\vspace{-.3cm}
Our final experiment is to apply CGMMN to distill knowledge from Bayesian models by learning a conditional distribution model for efficient prediction.
Specifically, let $\thetav$ denote the random variables.
A Bayesian model first computes the posterior distribution given the training set $\data = \{(\xv_i, \yv_i)\}_{i=1}^N$ as
$p(\thetav | \data)$. In the prediction stage, given a new input $\xv$, a response sample $\yv$ is generated
via probability $p(\yv|\xv,\data) = \int p(\yv|\xv,\thetav)p(\thetav|\data) d \thetav $.
This procedure usually involves a complicated integral thus is time consuming.
\cite{BDK:arXiv} show that we can learn a relatively simple {\it student network} to distill knowledge from the {\it teacher network} (i.e., the Bayesian model) and approximately represent the predictive distribution $p(\yv | \xv, \data)$ of the teacher network.

Our CGMMN provides a new solution to build such a student network for Bayesian dark knowledge.
To learn CGMMN, we need two datasets to estimate the CMMD objective --- one is generated by the teacher network and the other one is generated by CGMMN.
The former sampled dataset serves as the training
dataset for our CGMMN and the latter one is generated
during the training process of it.
For high-dimensional data, adopting the same strategy as~\cite{BDK:arXiv}, we sample ``near" the training data to generate the former dataset (i.e., perturbing the inputs in the training set slightly before sending to the teacher network to sample $\yv$).

Due to the space limitation, we test our model on a regression problem on the Boston housing dataset, which was also used in~\cite{BDK:arXiv,PBP:icml2015}, while deferring the other results on a synthetic dataset to Appendix~\ref{toy_bayes}.
The dataset consists of $506$ data points where each data is of dimension $13$. We
first train a PBP model~\cite{PBP:icml2015}, which is a scalable method for posterior inference in Bayesian neural networks, as the teacher and then distill it using our CGMMN
model. We test whether the distilled model will degrade the prediction performance.

\begin{wraptable}{r}{0.5\textwidth}\vspace{-.6cm}
\centering
\caption{Distilling results on Boston Housing dataset, the error is measured by RMSE}
\label{distill_bouston}
\begin{tabular}{c|c}
    \hline
    \hline
    PBP prediction  & Distilled by CGMMN \\
    \hline
    $2.574 \pm 0.089$ & $ 2.580 \pm 0.093$ \\
    \hline
\end{tabular}\vspace{-.3cm}
\end{wraptable}

We distill the PBP model~\cite{PBP:icml2015} using an MLP network with three hidden layers and $(100,50,50)$ hidden units for middle layers.
We draw $N = 3,000$ sample pairs $\{(\bm{x}_i,y_i)\}_{i=1}^{N}$
from the PBP network, where $\xv_i$ is the input variables that serve as conditional variables in
our model. For a fair comparison, $\bm{x}_i$ is generated by adding noise into training data to avoid
fitting the testing data directly. We evaluate the prediction performance on the original testing data by
root mean square error (RMSE).
Table \ref{distill_bouston} shows the results. We can see that the distilled model does not harm the prediction performance.
It is worth mentioning that we are not restricted to distill knowledge from PBP. In fact, any
Bayesian models can be distilled using CGMMN.

\vspace{-.3cm}
\section{Conclusions and Discussions}
\vspace{-.3cm}

We present conditional generative moment-matching networks (CGMMN),
which is a flexible framework to represent conditional distributions. CGMMN largely
extends the ability of previous DGM based on maximum mean discrepancy (MMD)
while keeping the training process simple as well, which is done by back-propagation.
Experimental results on various tasks, including predictive modeling, data generation and Bayesian dark knowledge, demonstrate competitive performance.

Conditional modeling has been practiced as a natural step towards improving the discriminative ability of a statistical model and/or relaxing unnecessary assumptions of the conditional variables. For deep learning models,
sum product networks (SPN)~\cite{SPN:uai11} provide exact inference on DGMs and its conditional extension~\cite{DSPN:nips12} improves the discriminative ability; and
the recent work~\cite{Mirza:arxiv14} presents a conditional version of the generative adversarial networks (GAN)~\cite{GAN:nips2014} with wider applicability.
Besides, the recent proposed conditional variational autoencoder~\cite{LSOR}
also works well on structured prediction.
Our work fills the research void to significantly improve the applicability of moment-matching networks.

\subsubsection*{Acknowledgments}
The work was supported by the National Basic Research Program (973 Program) of China (Nos.
2013CB329403, 2012CB316301), National NSF of China (Nos. 61322308, 61332007), Tsinghua TNList Lab
Big Data Initiative, and Tsinghua Initiative Scientific Research Program (Nos. 20121088071, 20141080934).

\small
\bibliography{report}

\begin{thebibliography}{10}

\bibitem{LapGAN:nips15}
E.~Denton, S.~Chintala, A.~Szlam, and R.~Fergus.
\newblock Deep generative image models using a laplacian pyramid of adversarial
  networks.
\newblock {\em NIPS}, 2015.

\bibitem{text2image}
A.~Dosovitskiy, J.~Springenberg, M.~Tatarchenko, and T.~Brox.
\newblock Learning to generate chairs, tables and cars with convolutional
  networks.
\newblock {\em arXiv:1411.5928}, 2015.

\bibitem{MMD:uai2015}
G.~Dziugaite, D.~Roy, and Z.~Ghahramani.
\newblock Training generative neural networks via maximum mean discrepancy
  optimization.
\newblock {\em UAI}, 2015.

\bibitem{DSPN:nips12}
R.~Gens and P.~Domingos.
\newblock Discriminative learning of sum-product networks.
\newblock {\em NIPS}, 2012.

\bibitem{GAN:nips2014}
I.~Goodfellow, J.~Pouget-Abadie, M.~Mirza, B.~Xu, D.~Warde-Farley, S.~Ozair,
  A.~Courville, and Y.~Bengio.
\newblock Generative adverisarial nets.
\newblock {\em NIPS}, 2014.

\bibitem{MN:icml2013}
I.~Goodfellow, D.~Warde-Farley, M.~Mirza, A.~Courville, and Y.~Bengio.
\newblock Maxout networks.
\newblock {\em ICML}, 2013.

\bibitem{KTST:jmlr}
A.~Gretton, K.~Borgwardt, M.~Rasch, B.~Scholkopf, and A.~Smola.
\newblock A kernel two-sample test.
\newblock {\em JMLR}, 2008.

\bibitem{CMMAR:icml2012}
S.~Grunewalder, G.~Lever, L.~Baldassarre, S.~Patterson, A.~Gretton, and
  M.~Pontil.
\newblock Conditional mean embedding as regressors.
\newblock {\em ICML}, 2012.

\bibitem{ext_yale_face}
X.~He, S.~Yan, Y.~Hu, P.~Niyogi, and H.~Zhang.
\newblock Face recognition using laplacianfaces.
\newblock {\em IEEE Trans. Pattern Anal. Mach. Intelligence}, 27(3):328--340,
  2005.

\bibitem{PBP:icml2015}
J.~Hernandez-Lobato and R.~Adams.
\newblock Probabilistic backpropagation for scalable learning of bayesian
  neural networks.
\newblock {\em ICML}, 2015.

\bibitem{kernelML2008}
T.~Hofmann, B.~Scholkopf, and A.~Smola.
\newblock Kernel methods in machine learning.
\newblock {\em The Annals of Statistics}, 36(3):1171--1220, 2008.

\bibitem{Prob}
O.~Kallenbery.
\newblock Foundations of modern probability.
\newblock {\em New York: Springer}, 2002.

\bibitem{Adam:iclr2015}
D.~Kingma and J.~Ba.
\newblock Adam: A method for stochastic optimization.
\newblock {\em ICLR}, 2015.

\bibitem{VA:iclr2014}
D.~Kingma and M.~Welling.
\newblock Auto-encoding variational bayes.
\newblock {\em ICLR}, 2014.

\bibitem{BDK:arXiv}
A.~Korattikara, V.~Rathod, K.~Murphy, and M.~Welling.
\newblock Bayesian dark knowledge.
\newblock {\em NIPS}, 2015.

\bibitem{crf2001}
J.~Lafferty, A.~McCallum, and F.~Pereira.
\newblock Conditional random fields: Probabilistic models for segmenting and
  labeling sequence data.
\newblock {\em ICML}, 2001.

\bibitem{DSN:aistats2015}
C.~Lee, S.~Xie, P.~Gallagher, Z.~Zhang, and Z.~Tu.
\newblock Deeply-supervised nets.
\newblock {\em AISTATS}, 2015.

\bibitem{MMDGM:nips15}
C.~Li, J.~Zhu, T.~Shi, and B.~Zhang.
\newblock Max-margin deep generative models.
\newblock {\em NIPS}, 2015.

\bibitem{GMMD:icml2015}
Y.~Li, K.~Swersky, and R.~Zemel.
\newblock Generative moment matching networks.
\newblock {\em ICML}, 2015.

\bibitem{NIN:iclr2014}
M.~Lin, Q.~Chen, and S.~Yan.
\newblock Network in network.
\newblock {\em ICLR}, 2014.

\bibitem{Mirza:arxiv14}
M.~Mirza and S.~Osindero.
\newblock Conditional generative adversarial nets.
\newblock {\em ArXiv:1411.1784v1}, 2014.

\bibitem{Relu:icml2010}
V.~Nair and G.~Hinton.
\newblock Rectified linear units improve restricted boltzmann machines.
\newblock {\em ICML}, 2010.

\bibitem{disc-vs-gen}
A.~Ng and M.I. Jordan.
\newblock On discriminative vs. generative classifiers: a comparison of
  logistic regression and naive bayes.
\newblock {\em NIPS}, 2001.

\bibitem{SPN:uai11}
H.~Poon and P.~Domingos.
\newblock Sum-product networks: A new deep architecture.
\newblock {\em UAI}, 2011.

\bibitem{CNN-SVHN:ICPR2012}
P.~Sermanet, S.~Chintala, and Y.~Lecun.
\newblock Convolutional neural networks applied to house numbers digit
  classification.
\newblock {\em ICPR}, 2012.

\bibitem{Pegasos}
S.~Shalev-Shwartz, Y.~Singer, N.~Srebro, and A.~Cotter.
\newblock Pegasos: Primal estimated sub-gradient solver for svm.
\newblock {\em Mathmetical Programming, Series B}, 2011.

\bibitem{KED:icalt2007}
A.~Smola, A.~Gretton, L.~Song, and B.~Scholkopf.
\newblock A hilbert space embedding for distributions.
\newblock {\em International Conference on Algorithmic Learning Theory}, 2007.

\bibitem{LSOR}
K.~Sohn, X.~Yan, and H.~Lee.
\newblock Learning structured output representation using deep conditional
  generative models.
\newblock {\em NIPS}, 2015.

\bibitem{HSECD:icml09}
L.~Song, J.~Huang, A.~Smola, and K.~Fukumizu.
\newblock Hilbert space embeddings of conditional distributions with
  applications to dynamical systems.
\newblock {\em ICML}, 2009.

\bibitem{multimodalDBM}
N.~Srivastava and R.~Salakhutdinov.
\newblock Multimodal learning with deep boltzmann machines.
\newblock {\em NIPS}, 2012.

\bibitem{image2text}
O.~Vinyals, A.~Toshev, S.~Bengio, and D.~Erhan.
\newblock Show and tell: A neural image caption generator.
\newblock {\em arXiv:1411.4555v2}, 2015.

\bibitem{attribute2image}
X.~Yan, J.~Yang, K.~Sohn, and H.~Lee.
\newblock Attribute2image: Conditional image generation from visual attributes.
\newblock {\em arXiv:1512.00570}, 2015.

\bibitem{SP:iclr2013}
M.~Zeiler and R.~Fergus.
\newblock Stochastic pooling for regularization of deep convolutional neural
  networks.
\newblock {\em ICLR}, 2013.

\end{thebibliography}
\bibliographystyle{plain}

\newpage
\appendix
\section{Appendix}
\normalsize

\subsection{Gradient Calculation}
\label{grad_cal}
The CMMD objective can be optimized by gradient descent. Precisely, for any network parameter
$\wv$, we have that:
\begin{eqnarray}
\dfrac{\partial \hat{\mathcal{L}}_{\cmmd}^2}{\partial \wv} =  \sum_{i=1}^M \dfrac{\partial \hat{\mathcal{L}}_{\cmmd}^2}{\partial \bm{y}^s_{i}} \dfrac{\partial \yv_i^s}{\partial \wv}, \nonumber
\end{eqnarray}
where the term $\dfrac{\partial \yv_i^s}{\partial \wv}$ can be calculated via back-propagation throughout the DNN and we use the chain rule to compute 
\begin{equation}
\begin{aligned}
	& \dfrac{\partial \hat{\mathcal{L}}_{\cmmd}^2}{\partial \bm{y}^s_{i}} = \mathrm{Tr}\left ( \widetilde{K}_s^{-1}
	K_s \widetilde{K}_s^{-1} \dfrac{\partial L_s}{\partial \bm{y}_i^s} \right )  -2 \cdot \mathrm{Tr} \left ( \widetilde{K}_s^{-1} K_{sd} \widetilde{K}_d^{-1}
	\dfrac{\partial L_{ds}}{\partial \bm{y}_i^s} \right ). \nonumber 
\end{aligned}
\end{equation}
The derivative of the kernel gram matrix (i.e., $\dfrac{\partial L_s}{\partial \yv_i^s}$ and $\dfrac{\partial L_{ds}}{\partial \yv_i^s}$)
can be calculated directly
as long as the kernel function of output samples $\yv$ is differentiable,
e.g., Gaussian RBF kernel $k_{\sigma}(\yv, \yv') = \exp\{- \frac{\|\yv - \yv'\|^2}{2\sigma^2}\}$.

\subsection{Implementation details}
Here we list some practical considerations to improve the performance of our models.
\vspace{-.1cm}
\subsubsection{Minibatch Training}
\label{mini-train}
\vspace{-.1cm}

The CMMD objective and its gradient involve an inverse operation on matrix such as $K_d + \lambda I$, which
has $O(N^3)$ time complexity to compute. This is unbearable when the data size is large.
Here, we present a minibatch based training algorithm to learn the CGMMN models. Specifically, in each training epoch, we first choose a
small subset $\mathcal B \subset \data$ and generate an equal number of samples based on the observation $\xv \in \mathcal B$
(i.e., we provide each $\xv \in \mathcal B$ to the generator to get a new sample).
The overall algorithm is provided in Alg.~$\ref{alg1}$. To further accelerate the
algorithm, we can pre-compute the inverse matrices $\widetilde{K}_d^{-1}$
and $\widetilde{K}_s^{-1}$ as cached data.

Essentially, the algorithm uses a single mini-batch to
approximate the whole dataset. When the dataset is ``simple" such as MNIST, a mini-batch of size
$200$ is enough to represent the whole dataset, however, for more complex datasets, larger mini-bath
size is needed.

\vspace{-.1cm}
\subsubsection{Finite Case for Conditional Variables}
\label{discrete}
\vspace{-.1cm}

Recall the empirical estimator of conditional kernel embedding operator as mentioned in Sec.
\ref{kecd}:
$
\widehat{C}_{Y|X} = \Phi(K+\lambda I)^{-1}\Upsilon^{\top},
$
where we need to compute the inverse of kernel gram matrix of the condition variables. Since
the domain of the variables is finite, the gram matrix is not invertible
in most cases. Although we can
impose a $\lambda$ to make the gram matrix invertible forcibly, this method cannot get the best result in
practice. Besides, the main effect of $\lambda$ is serving as regularization to avoid overfitting, not to make
the gram matrix invertible~\cite{CMMAR:icml2012}.

Fortunately, the problem can be avoided by choosing special kernels and estimating the conditional operator
directly. More precisely, we use Kronechker Delta kernel on conditioned variables $X$, i.e., $k(x,x') = \delta(x,x')$.
Suppose that $x \in \{1,...,K \}$, then the corresponding feature map $\phi(x)$ is the standard basis of $e_x \in \mathbb{R}^K$.
Recall that $C_{Y|X} = C_{YX}C_{XX}^{-1}$, instead of using the estimation before, we now can estimate $C_{XX}^{-1}$
directly since it can be expressed as follows:
$$
C_{XX}^{-1} = \left ( \begin{matrix}
	P(x=1)& ... &  0  \\
			 & \ddots & \\
	0 &... & p(x=K) \\
\end{matrix} \right ) ^{-1}.
$$
Obviously, the problem of inverse operator disappears.

\subsubsection{Kernel Choosing}
In general, we adopted Gaussian kernels as in GMMN. We also tried the strategy that combines several Gaussian kernels with different bandwidths, but it didn't make noticeable difference.

We tuned the bandwidth on the training set, and found that the bandwidth is appropriate if the distance of two samples (i.e., $\|x-y\|^2/\sigma^2$) is in range $[0,1]$.

\subsection{Distill Knowledge from Bayesian Models}
\label{toy_bayes}
We evaluate our model on a toy dataset, following the setting in~\cite{BDK:arXiv}. Specifically, the dataset is generated by random sampling
$20$ one-dimensional inputs $x$ uniformly in the interval $[-4,4]$. For each $x$, the response
variable $y$ is computed as $y = x^3 + \epsilon$, where $\epsilon \sim \mathcal{N}(0,9)$.

We first fit the data using probabilistic backpropagation (PBP)~\cite{PBP:icml2015}, which is a scalable method for
posterior inference in Bayesian neural networks. Then we use CGMMN with a two-layer MLP architecture, which is of size $(100,50)$,
to distill the knowledge for the PBP network (same architecture as CGMMN) using $3,000$ samples that are generated from it.

\begin{figure}[ht!]\vspace{-.5cm}
\centering
\subfigure[PBP prediction]{\includegraphics[width=6.2cm]{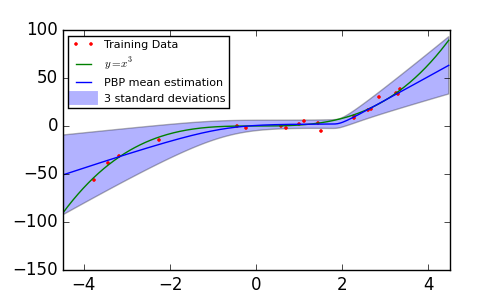}}
\subfigure[Distilled prediction]{\includegraphics[width=6.2cm]{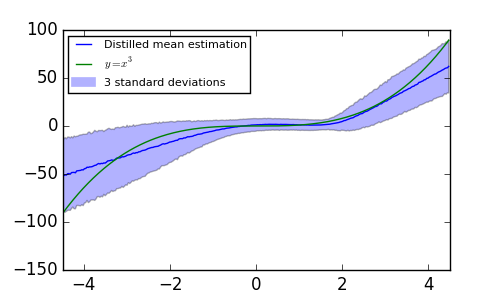}}\vspace{-.3cm}
\caption{Distilling results on toy dataset. (a) is the prediction given by PBP; (b) is the
distilled results using our model}
\label{distill_toy}\vspace{-.1cm}
\end{figure}
Fig.~\ref{distill_toy} shows the distilled results. We can see that the distilled model
is highly similar with the original one, especially on the mean estimation.

\subsection{More Results on Yale Face Dataset}

\subsubsection{Interpolation for Extended Yale Face samples}
\label{yale-inter}
\begin{wrapfigure}{r}{0.5\textwidth}\vspace{-0.65cm}
\centering
\includegraphics[width=5.2cm]{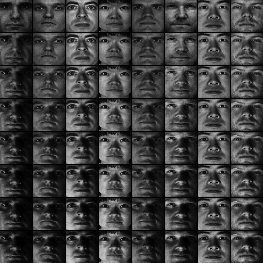}
\caption{Linear interpolation for Extended Yale Face Dataset.
Columns are conditioned on different individuals.}
\label{fig_yale_inter}\vspace{-0.7cm}
\end{wrapfigure}

One of the interesting aspects of a deep generative model
is that it is possible to directly explore the data manifold.
As well as to verify that our CGMMN will not merely copy the training data,
we perform linear interpolation on the first dimension of the hidden variables and set the
other dimensions to be $0$.
Here we use the same settings as in Sec.~\ref{yale_exp}.

Fig.~\ref{fig_yale_inter} shows the result. Each column is conditioned on a different
individual and we can find that for each individual, as the value of the first dimension varies, the
generated samples have the same varying trend in a continuous manner. This result verifies that
our model has a good latent representation for the training data and will not merely copy the
training dataset.

\newpage
\subsubsection{Results for smaller version of Yale Face Dataset}
\label{yale-small}
\begin{wrapfigure}{r}{0.66\textwidth}\vspace{-0.5cm}
\centering
\subfigure[Different individuals]{\includegraphics[width=4.5cm]{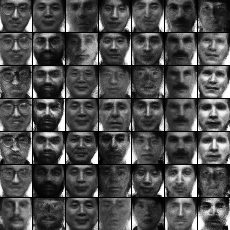}}
\subfigure[Individual $15$]{\includegraphics[width=4.5cm]{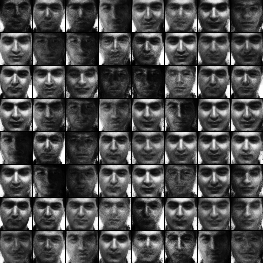}}
\caption{CGMMN generated samples for Yale Face Dataset. Columns in (a) are conditioned on different individuals while the label is $15$ in (b).}
\label{fig_yale}\vspace{-0.2cm}
\end{wrapfigure}
Here we show the generating result for the small version of Yale Face Dataset, which
consists of $165$ figures of $15$ individuals.
We adopt the same architecture as the generating experiments for MNIST, which
is a $5$-layer MLP with input dimension $30$ ($15$ hidden variable and $15$ dimension for
conditional variable) and the middle layer hidden unit number
$(64, 256, 256, 512)$. Since the dataset is small, we can run our algorithm with the whole
dataset as a mini-batch. The overall results are shown in Fig.~\ref{fig_yale}. We really
generate a wide diversity of different individuals. Obviously, our CGMMN will not merely copy the
training dataset since each figure of (b) in Fig.~\ref{fig_yale} is meaningful and unique.

\end{document}